\documentclass[10pt, a4paper]{article}

\usepackage{lrec-coling2024} 
\usepackage{todonotes}
\usepackage{amsfonts}
\usepackage{amsmath}
\usepackage{multirow}
\usepackage{adjustbox}
\usepackage{booktabs}
\usepackage{arydshln}
\usepackage{lipsum}
\usepackage{xcolor}

\title{Czech Dataset for Complex Aspect-Based Sentiment Analysis Tasks}

\name{Jakub \v{S}m\'{i}d\textsuperscript{*}, Pavel P\v{r}ib\'{a}\v{n}\textsuperscript{*}, Ond\v{r}ej Pra\v{z}\'{a}k\textsuperscript{*, $\dagger$}, Pavel Kr\'{a}l\textsuperscript{*, $\dagger$}} 

\address{University of West Bohemia, Faculty of Applied Sciences, \\
        \textsuperscript{*}Department of Computer Science and Engineering, \\
        \textsuperscript{$\dagger$}NTIS – New Technologies for the Information Society\\
         Univerzitní 2732/8, 301 00 Pilsen, Czech Republic \\
         \{jaksmid, pribanp, ondfa, pkral\}@kiv.zcu.cz\\
         \url{https://nlp.kiv.zcu.cz}
         }

\usepackage{fancyhdr}

\fancypagestyle{firstpage}{
  \fancyhf{} 
  \fancyfoot[C]{\thepage} 
  \fancyfoot[L]{\scriptsize Accepted at LREC-COLING 2024. Official version: \href{https://aclanthology.org/2024.lrec-main.384/}{aclanthology.org/2024.lrec-main.384/}} 
}

\makeatletter
\def\adl@drawiv#1#2#3{%
        \hskip.5\tabcolsep
        \xleaders#3{#2.5\@tempdimb #1{1}#2.5\@tempdimb}%
                #2\z@ plus1fil minus1fil\relax
        \hskip.5\tabcolsep}
\newcommand{\cdashlinelr}[1]{%
  \noalign{\vskip\aboverulesep
           \global\let\@dashdrawstore\adl@draw
           \global\let\adl@draw\adl@drawiv}
  \cdashline{#1}
  \noalign{\global\let\adl@draw\@dashdrawstore
           \vskip\belowrulesep}}
\makeatother

\newcommand{\quotes}[1]{``#1''}
\renewcommand{\vec}[1]{\ensuremath{\mathbf{#1}}}

\abstract{
In this paper, we introduce a novel Czech dataset for aspect-based sentiment analysis (ABSA), which consists of 3.1K manually annotated reviews from the restaurant domain. The dataset is built upon the older Czech dataset, which contained only separate labels for the basic ABSA tasks such as aspect term extraction or aspect polarity detection. Unlike its predecessor, our new dataset is specifically designed for more complex tasks, e.g. target-aspect-category detection. These advanced tasks require a unified annotation format, seamlessly linking sentiment elements (labels) together. Our dataset follows the format of the well-known SemEval-2016 datasets. This design choice allows effortless application and evaluation in cross-lingual scenarios, ultimately fostering cross-language comparisons with equivalent counterpart datasets in other languages. The annotation process engaged two trained annotators, yielding an impressive inter-annotator agreement rate of approximately 90\%.
Additionally, we provide 24M reviews without annotations suitable for unsupervised learning. We present robust monolingual baseline results achieved with various Transformer-based models and insightful error analysis to supplement our contributions. Our code and dataset are freely available for non-commercial research purposes.
 \\ \newline  }

\begin{document}

\maketitleabstract

\thispagestyle{firstpage} 

\section{Introduction}
Sentiment analysis (SA) is a widely recognized and fundamental field of 
natural language processing
that aims to understand and identify subjective information in text \citep{liu2012sentiment}. Sentiment classification (SC), known as polarity detection, is a common task within sentiment analysis that aims to classify a given text into one of pre-defined categories, such as \textit{positive}, \textit{negative} or \textit{neutral}.
\par
Aspect-based sentiment analysis (ABSA) is a more fine-grained task than SC. ABSA focuses on extracting detailed information about entities, their aspects and opinions expressed regarding these aspects. ABSA generally aims to identify the sentiment associated with each aspect or characteristic of a product or service. 
For instance, in restaurant reviews, opinions are not limited to the overall food quality but extend to other aspects like service, location, and atmosphere.
ABSA includes sentiment elements \citep{absa}, such as aspect term ($a$), aspect category ($c$), and sentiment polarity ($p$). In the review ($s$): \textit{\quotes{Delicious steak}}, these elements are \textit{\quotes{steak}}, \textit{food quality}, and \textit{positive}, respectively.
\par
ABSA involves several tasks \citep{absa}. Initially, research focused on identifying each sentiment element separately, such as aspect term extraction (ATE) or aspect category detection (ACD) \citep{pontiki-etal-2014-semeval}.
Recently, the focus has shifted to tasks that require linking sentiment elements in annotations, such as aspect polarity detection (APD). This linking also allows to predict more sentiment elements together, such as aspect-category-term extraction (ACTE) \citep{pontiki-etal-2015-semeval}, unified end-to-end ABSA (E2E-ABSA) \citep{wang2018towards}, and target-aspect-category detection (TASD) \citep{tasd}. Table \ref{tab:absa-tasks} shows input and output examples for selected ABSA tasks.

\begin{table}[ht!]
    \begin{adjustbox}{width=1.0\linewidth, center}
        \begin{tabular}{@{}llll@{}}
            \toprule
            \textbf{Task} & \textbf{Input}   & \textbf{Output}     & \textbf{Example output}                             \\ \midrule
            ATE           & $s$              & \{$a$\}             & \{\quotes{steak}, \quotes{water}\}                             \\
            ACD           & $s$              & \{$c$\}             & \{food, drinks\}                             \\
            APD           & $s$, \quotes{steak}, food & $p$                 & POS                                          \\
            E2E-ABSA      & $s$              & \{($a$, $p$)\}      & \{(\quotes{steak}, POS), (\quotes{water}, NEG)\}               \\
            ACTE          & $s$              & \{($a$, $c$)\}      & \{(\quotes{steak}, food), (\quotes{water}, drinks)\}           \\
            \multirow{2}{*}{TASD}          & \multirow{2}{*}{$s$}              & \multirow{2}{*}{\{($a$, $c$, $p$)\}} & \{(\quotes{steak}, food, POS),
            
            \\
            & & &(\quotes{water}, drinks, NEG)\} \\ \bottomrule
        \end{tabular}
    \end{adjustbox}
    \caption{Input and output format for ABSA tasks for a review $s$: \textit{\quotes{Delicious steak but expensive water}}.}
	\label{tab:absa-tasks}
\end{table}

\par For ABSA, several datasets have been built over time, including SemEval-2014 \citep{pontiki-etal-2014-semeval}, SemEval-2015 \citep{pontiki-etal-2015-semeval} and SemEval-2016 \citep{pontiki-etal-2016-semeval} datasets, SentiHood \citep{saeidi-etal-2016-sentihood} or Japanese dataset introduced by \citet{nakayama-etal-2022-large}. The datasets are mainly created for the English language except for the SemEval-2016, which also contains Arabic, Chinese, Dutch, French, Russian, Spanish and Turkish annotations. \citet{fan-etal-2019-target} and \citet{zhang-etal-2021-aspect-sentiment} provide datasets with opinion terms annotations. \citet{steinberger-etal-2014-aspect} and \citet{hercig2016unsupervised} introduced Czech datasets in the same format as the SemEval-2014 dataset, allowing a separate evaluation of the ATE, ACD and sentiment classification tasks in the Czech language. Similarly, \citet{tamchyna2015czech} presented a dataset with IT product reviews but annotated it only with a global review sentiment and aspect terms.

\par Unfortunately, the mentioned Czech datasets do not link the aspect term and category annotations, making it impossible to solve tasks where these two sentiment elements are predicted together, namely the TASD, ACTE and APD tasks. Therefore, the primary motivation of this paper is to provide a dataset that enables the evaluation of these advanced tasks in Czech. Consequently, the new dataset will allow cross-language comparison.

\par This paper presents a new dataset of 3,189 restaurant reviews tailored for complex ABSA tasks such as TASD and ACTE, which require annotations in a unified format linking individual labels together. Additionally, we crawled a set of 24M raw reviews intended for unsupervised learning. We reannotated the existing Czech dataset \citep{hercig2016unsupervised} and expanded it with more than 1,000 new reviews. The dataset adheres to the SemEval-2016 format, allowing evaluation of the more complex tasks and well-established existing tasks of ABSA in Czech. We describe the process of dataset creation and annotation. Two trained annotators annotated the dataset with an inter-annotator agreement of approximately 90\%.

\par We conduct a series of experiments and present robust baseline results utilizing Transformer-based models for the older ATE and ACD tasks, achieving 83.5\% and 85.7\% of the F1-score, respectively. Furthermore, we report baseline results for the complex APD, E2E, ACTE, and TASD tasks with 91.4\%, 75.5\%, 67.3\%, and 59.3\%, respectively. Finally, we provide an error analysis of sequence-to-sequence models, showing their weaknesses and limitations.

\par Our main contributions are the following: 1) We introduce a new Czech ABSA dataset\footnote{\label{foot:github}The annotated dataset, including its training splits, and code are freely available for research purposes at \url{https://github.com/biba10/Czech-Dataset-for-Complex-ABSA-Tasks}.} in the restaurant domain that allows solving more complex ABSA tasks and cross-lingual comparisons with SemEval-2016 datasets. 2) We perform experiments with Transformer-based models and provide robust baseline results with an error analysis.

\section{Related Work}
This section is devoted to existing and, according to our judgment, important ABSA datasets. Further, we review prior and recent works focused on aspect-based sentiment analysis, especially in Czech.

\subsection{ABSA Datasets}
Several ABSA datasets have been proposed. The SemEval-2014 dataset \citep{pontiki-etal-2014-semeval} contains English reviews from restaurants and laptop domains. The SemEval-2015 dataset \citep{pontiki-etal-2015-semeval} is based on the SemEval-2014 dataset with a more unified annotation format that links sentiment elements together. The SemEval-2016 dataset \citep{pontiki-etal-2016-semeval} is extended to new domains and provides more languages besides English, namely Arabic, Chinese, Dutch, French, Russian, Spanish and Turkish. These datasets lack the annotation of opinion terms. \citet{fan-etal-2019-target} provide the dataset with annotated opinion terms for English. \citet{zhang-etal-2021-aspect-sentiment} introduce English datasets for different domains containing the annotations of four sentiment elements. The MAMS dataset \citep{jiang-etal-2019-challenge} is another dataset that focuses on the restaurant domain. Twitter is another valuable linguistic data resource, and \citet{dong-etal-2014-adaptive} constructed a dataset from Twitter comments. The SentiHood dataset \citep{saeidi-etal-2016-sentihood} is derived from a question-answering platform where users discuss urban neighbourhoods. \citet{nakayama-etal-2022-large} introduced the ABSA dataset for Japanese and \citet{hyun-etal-2020-building} for Korean.

\par

Compared to English, to the best of our knowledge, no ABSA dataset for Czech could be used for compound ABSA tasks, e.g. the TASD task. \citet{steinberger-etal-2014-aspect} introduced the first Czech ABSA dataset based on data from restaurant reviews, with the same type of annotations as in \citep{pontiki-etal-2014-semeval}. \citet{tamchyna2015czech} provide a dataset containing reviews of IT products with aspect term and sentiment polarity annotations. Unlike the Czech restaurant dataset by \citet{steinberger-etal-2014-aspect}, the IT product reviews are annotated with overall sentiment and aspect terms but lack categorization and sentiment classification for these terms. \citet{hercig2016unsupervised} expanded the Czech restaurant review ABSA dataset. The annotation format of the Czech restaurant datasets is based on the SemEval-2014 dataset and lacks linking aspect terms and categories. Moreover, these datasets have fewer, less detailed categories. For example, there is only a simple \textit{food} category, and the \textit{DRINKS} category is absent. However, the SemEval-2015 and SemEval-2016 datasets use categories in \textit{E\#A} format, combining entities and attributes, e.g. \textit{FOOD\#QUALITY} or \textit{FOOD\#PRICES}. See Figure \ref{fig:annotationexample} for an example.

\subsection{Aspect-Based Sentiment Analysis}
In recent years, there has been relatively little research on the ABSA task in Czech, and the existing approaches are often outdated compared to more modern sentiment classification methods. The pioneering work on Czech ABSA was done by \citet{steinberger-etal-2014-aspect}, who presented baseline results for their restaurant dataset using Conditional Random Fields (CRF) and Maximum Entropy (ME) classifiers. \citet{tamchyna2015czech} provided baseline results using CRF on their IT products dataset. Furthermore, \citet{hercig2016unsupervised} proposed various unsupervised methods to enhance performance in ABSA tasks for Czech and English, utilizing CRF and ME classifiers. Their research demonstrated that unsupervised methods could yield significant performance improvements. \citet{SLON-Lenc2016Neural} employed a convolutional neural network (CNN) and recurrent neural network (RNN) for the task of identifying the sentiment polarity of aspect categories, evaluating their proposed model on the dataset from \citet{hercig2016unsupervised}.

\textcolor{black}{Most recently, \citet{smid-priban-2023-prompt} introduced the first prompt-based approach for SC and ABSA in Czech using models based on the Transformer architecture \citep{vaswani2017attention}. One of their methods can solve multiple ABSA tasks simultaneously using sequence-to-sequence models. They also show the effectiveness of prompting in few-shot settings and that in-domain pre-training improves the results. \citet{priban-prazak-2023-improving} combined ABSA tasks with the semantic information obtained by solving the semantic role labelling task. The multitask combination effectively improved results for Czech and English ACD and category polarity tasks.}

To address the relative lack of work for Czech ABSA and provide an overview of the latest approaches, we present example studies focusing on ABSA in English, the most studied language in this field. \citet{li-etal-2019-exploiting} demonstrated the effectiveness of a BERT-based architecture for the E2E-ABSA task. Recent ABSA research primarily treats it as a text generation problem. \citet{zhang-etal-2021-towards-generative} introduced two paradigms for ABSA tasks designed to produce text output in a desired format, employing the English T5 model \citep{raffel2020exploring}. They achieved new state-of-the-art (SotA) benchmarks across various ABSA tasks, including TASD, using datasets from the SemEval competitions \citep{pontiki-etal-2014-semeval, pontiki-etal-2015-semeval, pontiki-etal-2016-semeval}. Similarly, \citet{zhang-etal-2021-aspect-sentiment} utilized the same English T5 model to address a recently introduced ABSA task known as aspect sentiment quad prediction, generating text output as well. 
\citet{gou-etal-2023-mvp} introduced a method combining outputs with different orders of sentiment elements, showing that the order of elements matters, and achieved new SotA results on different datasets.

\section{Dataset Construction}
This paper aims to build a Czech dataset for ABSA within the restaurant domain, utilizing the annotation format consistent with SemEval-2015 \citep{pontiki-etal-2015-semeval} and SemEval-2016 \citep{pontiki-etal-2016-semeval} datasets. This annotation format is instrumental in addressing more complex ABSA tasks, including APD, ACTE and TASD. Additionally, aligning the annotation format with SemEval-2016 allows future cross-lingual experiments between Czech, English, and other languages in the SemEval-2016 dataset.

\par Our newly created dataset consists of two parts: a) manually annotated 3,189 reviews, see Tables \ref{tab:data_stats} and \ref{tab:detailed_stats} for detailed statistics and b) 24M raw additional reviews (330 MB of plain text) intended for additional unsupervised learning.

\subsection{Unsupervised Dataset}
\par The unsupervised part of the dataset was crawled from restaurant reviews on Google Maps\footnote{\url{http://googlemaps.com/}}. Firstly, we obtained the list of names of Czech restaurants from the Restu.cz\footnote{\url{https://restu.cz/}} website. Consequently, we searched each of the obtained restaurant names with Google Maps and downloaded all available reviews for the particular restaurant during September 2022. To maintain a certain level of anonymity, we provide only the reviews in the dataset. Additional details like the restaurant name, review date, or author's name are not included.

\subsection{Manual Annotation}
The manually annotated part of the dataset comprises two data segments. Firstly, we reuse all the reviews from the original dataset from \citet{hercig2016unsupervised}. Secondly, we randomly sampled 1,110 reviews from the unsupervised part of the dataset. The existing annotations in the SemEval-2014 format cannot be directly converted into the SemEval-2016 format, and all the reviews must be read and labelled again from scratch. Thus, we started by completely reannotating the dataset from \citet{hercig2016unsupervised} following the annotations guidelines provided by \citet{pontiki-etal-2015-semeval}. Two native speakers annotated all the original data. For a given review, each annotator had the following tasks:
\begin{enumerate}
    \itemsep0em 
    \item \textbf{Identify objective reviews:} Objective reviews and reviews without any sentiment expressed had to be marked\footnote{To allow potential comparison and to keep the backward compatibility and consistency with the original dataset, we did not exclude the objective reviews.} as \textit{\quotes{OutOfScope}}. Example: \textit{\quotes{Koupila jsem 3 vouchery na pizzu.} (\quotes{I bought 3 pizza vouchers.})}.

    \item \textbf{Identify aspect terms:} One or more word expressions naming a specific aspect of the target entity, e.g. \textit{\quotes{toast s vejci} (\quotes{toast with eggs})}. Implicitly referred aspect terms, e.g. in a review \textit{\quotes{Doporučuji} (\quotes{Recommended})}, had to be assigned the value \quotes{NULL}.
    
    \item \textbf{Assign aspect category:} The annotators had to assign aspect categories for each identified aspect term. The aspect category consists of entity and attribute (\textit{E\#A}) and must be chosen from a pre-defined set of categories (e.g. \textit{RESTAURANT\#GENERAL} or \textit{FOOD\#QUALITY}). One aspect term could be assigned more aspect categories (for example, if the review mentions the quality and cost of the same aspect term). Example: \textit{\quotes{Rychlá obsluha} (\quotes{Fast service}) – \quotes{obsluha} (\quotes{service}) $\rightarrow$ SERVICE\#GENERAL}.
    
    \item \textbf{Assign sentiment polarity:} Finally, for each (aspect term, aspect category) tuple, the annotators had to assign the sentiment polarity from one of the following values: \textit{neutral}, \textit{negative}, and \textit{positive}. The \textit{neutral} polarity applies to mildly negative or mildly positive sentiment.
\end{enumerate}

\begin{figure}[ht!]
    \centering
    \includegraphics[scale=1.25]{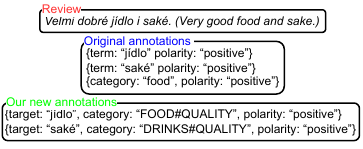}
    \caption{Our new annotations for a sample review compared to annotations from \citet{hercig2016unsupervised}.}
    \label{fig:annotationexample}
\end{figure}

An example of a review with annotation triplets compared to the annotations format used by \citet{hercig2016unsupervised} is shown in Figure \ref{fig:annotationexample}.

\subsection{Annotators Details}
\textcolor{black}{Before the annotation of our dataset, all annotators have thoroughly passed the guidelines for annotations for SemEval 2015 \citep{pontiki-etal-2015-semeval} and SemEval 2016 \citep{pontiki-etal-2016-semeval} datasets and made a shared document with the important points regarding the annotation. Additionally, all annotators went through a few hundredths of annotated reviews from the English restaurant dataset from SemEval 2016 and made additional comments to the shared document. Then, all annotators discussed the main points.}

\par \textcolor{black}{Subsequent to the initial discussion, the annotators started the annotation process. During the annotation, after every few hundred new annotated data, the annotators reviewed the problems (if any) and went through the comments again. This procedure ensured the best possible annotator agreement and mitigated a lot of potential issues that might have occurred. Therefore, we only encountered those mentioned in Section \ref{sec:inter-annotation-agreement}}.

\subsection{Inter-annotator Agreement}
\label{sec:inter-annotation-agreement}
\par Following \citet{pontiki-etal-2016-semeval, steinberger-etal-2014-aspect, hercig2016unsupervised}, we calculated the inter-annotator agreement (IAA) as F1-score, where annotations from one annotator are treated as gold data and annotation from the second annotator as predictions. Table \ref{tab:agreement} shows the agreement. Similarly, \citet{pontiki-etal-2016-semeval} achieved comparable results with values between 80 and 91\% of IAA for the Spanish dataset. This fact indicates a similar level of quality of our dataset.

\begin{table}[ht!]
    \centering
    \begin{adjustbox}{width=0.85\linewidth}
        \begin{tabular}{@{}lr@{}}
            \toprule
            \textbf{Annotation target}                 & \textbf{IAA} \\ \midrule
            Aspect term                                & 93.19        \\
            Aspect category                            & 93.00        \\
            Aspect term \& aspect category             & 91.06        \\
            Aspect term \& aspect category \& polarity & 89.70        \\ \bottomrule
        \end{tabular}
    \end{adjustbox}
    \caption{Inter-annotator agreement (IAA) for different annotation targets in the new Czech dataset, measured in terms of micro F1 score (in \%).}
    \label{tab:agreement}
\end{table}

\par The main reasons for disagreements were mainly in the sentiment polarity, where, in some cases, it is difficult to determine whether the polarity is slightly positive or negative, hence neutral, or whether it should be assigned as strongly negative or positive. The definition of \textit{RESTAURANT\#GENERAL} and \textit{RESTAURANT\#MISCELLANEOUS} categories in the guidelines and datasets provided by \citet{pontiki-etal-2015-semeval, pontiki-etal-2016-semeval} are ambiguously defined. These two categories were another primary source of disagreement. The additional third annotator resolved the disagreement cases. 

\par Following the approach described above, we additionally annotated 1,110 new examples of the randomly sampled reviews from the unsupervised part of the dataset. We then removed the randomly sampled reviews from the unsupervised dataset to avoid their potential use for training, as the newly annotated reviews may also be present in test data. We also removed all the original data we found in the unsupervised dataset.

\par Given that we considered the agreements substantially high for all annotation targets, we split all the new reviews not presented in the original datasets from \citet{hercig2016unsupervised} into two parts. Each annotator then independently annotated one part.
Following annotation, each aspect's starting and ending offsets were automatically generated and labelled as \quotes{from} and \quotes{to} in the dataset.

We filtered out reviews without opinion triplets (i.e. objective reviews) and reviews in languages other than Czech from the new part. Example of filtered objective review: \textit{\quotes{Bar navštěvovaný mladými lidmi} (\quotes{Bar frequented by young people})}. After this filtration, 1,040 reviews remained in the dataset.

\subsection{Dataset Details}

\par To enhance future research, we providethree versions of our dataset, named \texttt{CsRest}\footnote{The versions suffix names come from O -- \textbf{O}riginal, N -- \textbf{N}ew and M -- \textbf{M}ixed.}, with train, validation and test splits. The first version (\texttt{CsRest-O}) exclusively comprises the reannotated data from \citep{hercig2016unsupervised}, with 25\% of this data designated as the test set. The other versions (\texttt{CsRest-N} and \texttt{CsRest-M}) contain all data, including newly annotated data not present in the original dataset \citep{hercig2016unsupervised}. In \texttt{CsRest-N}, all the new data serves as the test data, while in \texttt{CsRest-M}, we joined all the data together, shuffled them, and randomly selected 25\% as the test data. 
For all three versions of the dataset, we further split the data not included in the test set into training and validation sets in a 9:1 ratio.
The selection of 25\% for test size is based on a similar value used in \citet{pontiki-etal-2016-semeval}.
\par
We significantly expanded the original dataset by almost 50\%, increasing the number of reviews from 2,149 to 3,189 for our dataset's \texttt{CsRest-N} and \texttt{CsRest-M} versions. This expansion introduced more than a 75\% growth in the number of triplets, from 3,670 to 6,478, compared to the \texttt{CsRest-O} version of our dataset, which exclusively contains data from the original dataset. Compared to the SemEval-2016 datasets, the two larger versions (\texttt{CsRest-N} and \texttt{CsRest-M}) of the Czech ABSA dataset now stand as the second largest restaurant domain datasets regarding the number of reviews (only behind the Russian version) and the largest in the number of annotation triplets.

\par
Table \ref{tab:data_stats} shows the statistics of the dataset in terms of the number of reviews, annotation triplets, reviews without any annotation triplets and the number of \quotes{NULL} aspect terms (i.e. implicitly mentioned). Table \ref{tab:detailed_stats} shows detailed statistics regarding aspect categories and sentiment polarity. We can see the imbalance in both aspect categories and sentiment polarity. The \textit{neutral} sentiment polarity is the least frequent. The reviews most often mention the food quality and the restaurant and service. On the other hand, they do not often mention the location or prices. \textcolor{black}{Table \ref{tab:data_stats_all} shows a comparison between our dataset and those in other languages within the restaurant domain in terms of the number of reviews and annotated triplets.}
\begin{table}[ht!]
    \centering
    \footnotesize
    \begin{adjustbox}{width=0.95\linewidth, center}
    \begin{tabular}{@{}llrrr@{}}

        \textbf{Split}         & \textbf{Count} & \textbf{\texttt{CsRest-O}} & \textbf{\texttt{CsRest-N}} & \textbf{\texttt{CsRest-M}} \\ \midrule
            \multirow{4}{*}{Train} & Reviews        & 1,450         & 1,934        & 2,151          \\
                                   & Triplets       & 2,510         & 3,240        & 4,386          \\
                                   & No triplets    & 104          & 142          & 109            \\
                                   & NULL terms     & 590          & 795          & 961            \\  \cdashlinelr{1-5}
            \multirow{4}{*}{Dev}   & Reviews        & 162          & 215          & 240            \\
                                   & Triplets       & 253          & 430          & 483            \\
                                   & No triplets    & 6            & 17           & 9              \\
                                   & NULL terms     & 64           & 95           & 104            \\  \cdashlinelr{1-5}
            \multirow{4}{*}{Test}  & Reviews        & 537          & 1,040        & 798            \\
                                   & Triplets       & 907          & 2,808        & 1,609          \\
                                   & No triplets    & 49           & 0            & 41             \\
                                   & NULL terms     & 49           & 517          & 342            \\ \midrule
            \multirow{4}{*}{Total} & Reviews        & 2,149         & 3,189        & 3,189          \\
                                   & Triplets       & 3,670         & 6,478        & 6,478          \\
                                   & No triplets    & 159          & 159          & 159            \\
                                   & NULL terms     & 890          & 1,407        & 1,407          \\ \bottomrule
    \end{tabular}
    \end{adjustbox}
    \caption{Statistics of our dataset.}
    \label{tab:data_stats}
\end{table}

\begin{table}[ht!]
    \centering
    \footnotesize
        \begin{adjustbox}{width=0.65\linewidth, center}
        \begin{tabular}{@{}crr@{}}
        \toprule
        \textbf{Dataset} & \textbf{Reviews} & \textbf{Triplets} \\ \midrule
        en               & 2,676            & 3,366             \\
        es               & 2,951            & 3,792             \\
        fr               & 2,429            & 5,322             \\
        nl               & 2,286            & 2,473             \\
        ru               & 4,699            & 5,322             \\
        tr               & 1,248            & 1,694             \\
        \cdashlinelr{1-3}
        CsRest-O         & 2,149            & 3,670             \\
        CsRest-N         & 3,189            & 6,478             \\
        CsRest-M         & 3,189            & 6,478             \\ \bottomrule
        \end{tabular}
    \end{adjustbox}
    \caption{Statistics of our dataset compared to datasets in another languages in the restaurant domain provided by \citet{pontiki-etal-2016-semeval}.}
    \label{tab:data_stats_all}
\end{table}

\par 

Our newly annotated dataset offers several improvements compared to the original dataset \citep{hercig2016unsupervised}. It links information between aspect terms and categories and aligns with the SemEval-2016 dataset, allowing us to perform more advanced tasks. It also provides more detailed annotations. For example, our annotations comprise entities and attributes in \textit{E\#A} format, e.g. \textit{FOOD\#QUALITY} or \textit{FOOD\#PRICES}, whereas the original dataset would merge them into a \textit{food} category. Additionally, our dataset introduces new categories (entities) not present in the original dataset, such as \quotes{DRINKS}.

\newcommand{\STAB}[1]{\begin{tabular}{@{}c@{}}#1\end{tabular}}

\begin{table*}[ht!]
    \centering
    \begin{adjustbox}{width=1\linewidth, center}
        \begin{tabular}{@{}clrrrrrrrrrrrrrrrr@{}}
            \toprule
            \multirow{2}{*}{} & \multirow{2}{*}{\textbf{Category}} & \multicolumn{4}{c}{\textbf{Train}} & \multicolumn{4}{c}{\textbf{Dev}} & \multicolumn{4}{c}{\textbf{Test}} & \multicolumn{4}{c}{\textbf{Total}} \\ \cmidrule(lr){3-6} \cmidrule(lr){7-10} \cmidrule(lr){11-14} \cmidrule(l){15-18} 
                                              &                                    & Pos     & Neg     & Neu   & Tot    & Pos    & Neg    & Neu    & Tot   & Pos     & Neg    & Neu   & Tot    & Pos     & Neg     & Neu   & Tot    \\ \midrule
            \multirow{13}{*}{\STAB{\rotatebox[origin=c]{90}{\texttt{CsRest-O}}}}             & AMBIENCE\#GENERAL                  & 89                      & 75   & 8   & 172  & 5    & 14   & 1   & 20  & 41   & 20   & 4    & 65  & 135   & 109  & 13  & 257  \\
                                  & DRINKS\#PRICES                     & 2                       & 10   & 8   & 20   & 0    & 0    & 0   & 0   & 0    & 1    & 0    & 1   & 2     & 11   & 8   & 21   \\
                                  & DRINKS\#QUALITY                    & 61                      & 30   & 16  & 107  & 4    & 6    & 1   & 11  & 22   & 14   & 2    & 38  & 87    & 50   & 19  & 156  \\
                                  & DRINKS\#STYLE\_OPTIONS             & 7                       & 13   & 6   & 26   & 0    & 2    & 1   & 3   & 1    & 0    & 0    & 1   & 8     & 15   & 7   & 30   \\
                                  & FOOD\#PRICES                       & 18                      & 38   & 11  & 67   & 0    & 3    & 2   & 5   & 2    & 16   & 3    & 21  & 20    & 57   & 16  & 93   \\
                                  & FOOD\#QUALITY                      & 400                     & 275  & 87  & 762  & 32   & 29   & 12  & 73  & 166  & 113  & 16   & 295 & 598   & 417  & 115 & 1,130 \\
                                  & FOOD\#STYLE\_OPTIONS               & 49                      & 84   & 41  & 174  & 5    & 4    & 2   & 11  & 10   & 33   & 7    & 50  & 64    & 121  & 50  & 235  \\
                                  & LOCATION\#GENERAL                  & 5                       & 4    & 0   & 9    & 2    & 0    & 0   & 2   & 2    & 0    & 1    & 3   & 9     & 4    & 1   & 14   \\
                                  & RESTAURANT\#GENERAL                & 278                     & 229  & 35  & 542  & 23   & 31   & 2   & 56  & 99   & 99   & 13   & 211 & 400   & 359  & 50  & 809  \\
                                  & RESTAURANT\#MISCELLANEOUS          & 5                       & 52   & 25  & 82   & 3    & 6    & 6   & 15  & 2    & 24   & 1    & 27  & 10    & 82   & 32  & 124  \\
                                  & RESTAURANT\#PRICES                 & 21                      & 27   & 9   & 57   & 1    & 4    & 1   & 6   & 13   & 14   & 4    & 31  & 35    & 45   & 14  & 94   \\
                                  & SERVICE\#GENERAL                   & 209                     & 231  & 52  & 492  & 12   & 30   & 9   & 51  & 62   & 92   & 10   & 164 & 283   & 353  & 71  & 707  \\ \cdashlinelr{2-18} 
                                  & Total                              & 1,144                    & 1,068 & 298 & 2,510 & 87   & 129  & 37  & 253 & 420  & 426  & 61   & 907 & 1,651  & 1,623 & 396 & 3,670 \\ \midrule
            \multirow{13}{*}{\STAB{\rotatebox[origin=c]{90}{\texttt{CsRest-N}}}}             & AMBIENCE\#GENERAL                  & 112     & 99      & 11    & 222    & 23     & 10     & 2      & 35    & 316     & 25     & 22    & 363    & 451     & 134     & 35    & 620    \\
                                  & DRINKS\#PRICES                     & 0       & 8       & 6     & 14     & 2      & 3      & 2      & 7     & 10      & 13     & 2     & 25     & 12      & 24      & 10    & 46     \\
                                  & DRINKS\#QUALITY                    & 78      & 48      & 17    & 143    & 9      & 2      & 2      & 13    & 179     & 20     & 8     & 207    & 266     & 70      & 27    & 363    \\
                                  & DRINKS\#STYLE\_OPTIONS             & 8       & 14      & 6     & 28     & 0      & 1      & 1      & 2     & 42      & 7      & 5     & 54     & 50      & 22      & 12    & 84     \\
                                  & FOOD\#PRICES                       & 19      & 50      & 13    & 82     & 1      & 7      & 3      & 11    & 36      & 26     & 15    & 77     & 56      & 83      & 31    & 170    \\
                                  & FOOD\#QUALITY                      & 527     & 373     & 103   & 1,003   & 71     & 44     & 12     & 127   & 698     & 94     & 48    & 840    & 1,296    & 511     & 163   & 1,970   \\
                                  & FOOD\#STYLE\_OPTIONS               & 54      & 112     & 42    & 208    & 10     & 9      & 8      & 27    & 108     & 27     & 8     & 143    & 172     & 148     & 58    & 378    \\
                                  & LOCATION\#GENERAL                  & 5       & 3       & 0     & 8      & 4      & 1      & 1      & 6     & 36      & 3      & 0     & 39     & 45      & 7       & 1     & 53     \\
                                  & RESTAURANT\#GENERAL                & 346     & 321     & 47    & 714    & 54     & 38     & 3      & 95    & 360     & 44     & 16    & 420    & 760     & 403     & 66    & 1,229   \\
                                  & RESTAURANT\#MISCELLANEOUS          & 8       & 77      & 28    & 113    & 2      & 5      & 4      & 11    & 22      & 13     & 4     & 39     & 32      & 95      & 36    & 163    \\
                                  & RESTAURANT\#PRICES                 & 32      & 40      & 13    & 85     & 3      & 5      & 1      & 9     & 51      & 20     & 21    & 92     & 86      & 65      & 35    & 186    \\
                                  & SERVICE\#GENERAL                   & 248     & 310     & 62    & 620    & 35     & 43     & 9      & 87    & 404     & 81     & 24    & 509    & 687     & 434     & 95    & 1,216   \\ \cdashlinelr{2-18} 
                                  & Total                              & 1,437    & 1,455    & 348   & 3,240   & 214    & 168    & 48     & 430   & 2,262    & 373    & 173   & 2,808   & 3,913    & 1,996    & 569   & 6,478   \\ \midrule
\multirow{13}{*}{\STAB{\rotatebox[origin=c]{90}{\texttt{CsRest-M}}}}           & AMBIENCE\#GENERAL                  & 306     & 90      & 26    & 422    & 35     & 7      & 3      & 45    & 110     & 37     & 6     & 153    & 451     & 134     & 35    & 620    \\
                                  & DRINKS\#PRICES                     & 8       & 13      & 5     & 26     & 1      & 1      & 4      & 6     & 3       & 10     & 1     & 14     & 12      & 24      & 10    & 46     \\
                                  & DRINKS\#QUALITY                    & 181     & 51      & 15    & 247    & 20     & 5      & 2      & 27    & 65      & 14     & 10    & 89     & 266     & 70      & 27    & 363    \\
                                  & DRINKS\#STYLE\_OPTIONS             & 32      & 19      & 12    & 63     & 2      & 2      & 0      & 4     & 16      & 1      & 0     & 17     & 50      & 22      & 12    & 84     \\
                                  & FOOD\#PRICES                       & 38      & 55      & 18    & 111    & 7      & 6      & 2      & 15    & 11      & 22     & 11    & 44     & 56      & 83      & 31    & 170    \\
                                  & FOOD\#QUALITY                      & 882     & 366     & 116   & 1,364   & 86     & 39     & 9      & 134   & 328     & 106    & 38    & 472    & 1,296    & 511     & 163   & 1,970   \\
                                  & FOOD\#STYLE\_OPTIONS               & 119     & 91      & 41    & 251    & 15     & 10     & 6      & 31    & 38      & 47     & 11    & 96     & 172     & 148     & 58    & 378    \\
                                  & LOCATION\#GENERAL                  & 28      & 1       & 1     & 30     & 1      & 2      & 0      & 3     & 16      & 4      & 0     & 20     & 45      & 7       & 1     & 53     \\
                                  & RESTAURANT\#GENERAL                & 525     & 264     & 44    & 833    & 55     & 38     & 5      & 98    & 180     & 101    & 17    & 298    & 760     & 403     & 66    & 1,229   \\
                                  & RESTAURANT\#MISCELLANEOUS          & 23      & 64      & 22    & 109    & 2      & 7      & 2      & 11    & 7       & 24     & 12    & 43     & 32      & 95      & 36    & 163    \\
                                  & RESTAURANT\#PRICES                 & 58      & 47      & 24    & 129    & 3      & 4      & 2      & 9     & 25      & 14     & 9     & 48     & 86      & 65      & 35    & 186    \\
                                  & SERVICE\#GENERAL                   & 463     & 277     & 61    & 801    & 51     & 40     & 9      & 100   & 173     & 117    & 25    & 315    & 687     & 434     & 95    & 1,216   \\ \cdashlinelr{2-18} 
                                  & Total                              & 2,663    & 1,338    & 385   & 4,386   & 278    & 161    & 44     & 483   & 972     & 497    & 140   & 1,609   & 3,913    & 1,996    & 569   & 6,478   \\ \bottomrule
        \end{tabular}
    \end{adjustbox}
    \caption{Detailed statistics of our dataset regarding aspect categories and sentiment polarity.}
    \label{tab:detailed_stats}
\end{table*}

\section{Experiments \& Setup}
To evaluate the quality of the proposed dataset, we conduct experiments on the following tasks:
\begin{itemize}
\itemsep-0.2em 
    \item Aspect term extraction (ATE): Extraction of aspect terms.
    \item Aspect category detection (ACD): Detection of aspect categories.
    \item Aspect-category-term extraction: Extraction of (aspect term, aspect category) tuples.
    \item Aspect polarity detection (APD): Detection of sentiment polarity for given (aspect term, aspect category) tuples.
    \item End-to-end ABSA (E2E-ABSA): Extraction of (aspect term, sentiment polarity) tuples.
    \item Target-aspect-sentiment detection (TASD): Detection of (aspect term, aspect category, sentiment polarity) triples.
\end{itemize}
For all tasks, we use the micro F1-score as evaluation metrics, and following related work \citep{zhang-etal-2021-aspect-sentiment, gou-etal-2023-mvp, zhang-etal-2021-towards-generative}, we discard all examples without any annotations (i.e. objective reviews).

\subsection{Encoder-Based Models}
We use encoder-based (BERT-like) models to perform ATE, ACD, E2E-ABSA and APD tasks. We employ four Czech pre-trained Transformer-based models, specifically Czert \citep{sido2021czert}, RobeCzech \citep{straka2021robeczech}, FERNET \citep{FERNET-2021} and Small-E-Czech \citep{kocian2022siamese}. Additionally, we use three multilingual models, specifically the multilingual BERT (mBERT) \citep{devlin-etal-2019-bert} and the base and large version of XLM-RoBERTa (XLM-R) \citep{conneau-etal-2020-unsupervised}. The encoder-based models convert an input sequence $x = w_1, \ldots, w_n$ of $n$ tokens into a sequence of hidden vectors $\vec{h} = \vec{h}_0, \vec{h}_1, \ldots, \vec{h}_n$. The hidden vector $\vec{h}_0 = \vec{h}_{\text{\texttt{[CLS]}}}$ is the artificial classification \texttt{[CLS]} token representing the entire input sequence. For each task, we utilize a linear layer on top of the model to generate predictions and fine-tune the model's parameters $\vec{\Theta}$ that include task-specific parameters $\vec{W}$ and $\vec{b}$.
\par
For the APD task, we create one input for each (aspect term, aspect category) tuple, where we append the tuple after the original review (the only task we solve where the number of inputs can be larger than the number of reviews). The linear layer computes the probability of a label $y$ from a label space $\mathcal{Y}\in\{\textit{positive}, \textit{negative}, \textit{neutral}\}$ for the input $x_i$ as
\begin{equation}
    P_\vec{\Theta}(y|x_i)={\text{softmax}}(\vec{W}\vec{h}_{\text{\texttt{[CLS]}}} + \vec{b}).
\end{equation}
We choose the class with the largest probability.
\par
The ACD task is similar to the APD task, but the label space is different; it contains all possible aspect categories. This task is also multi-label and not multi-class classification; hence, 0 to $k$ classes can be selected instead of precisely one, where $k$ is the total number of classes. We select all classes with a probability larger than 0.5.
\par
To each token, a label is assigned for the ATE and E2E-ABSA tasks using BIO tagging, which denotes the aspect boundaries. For the ATE task, the class $y_i$ for each token $x_i$ comes from a label space $\mathcal{Y}\in\{\textit{B}, \textit{I}, \textit{O}\}$, and for the E2E-ABSA task, from a label space $\mathcal{Y}\in\{\textit{B}, \textit{I}\}-{\textit{POS}, \textit{NEG}, \textit{NEU}}\cup \{\textit{O}\}$. For example, $y_i=\textit{B-NEG}$ means $x_i$ is the beginning of a negative aspect term. The label distribution of $x_i$ is computed as
\begin{equation}
    P_\vec{\Theta}(y_i|x_i)={\text{softmax}}(\vec{W}\vec{h}_i + \vec{b}).
\end{equation}
In the case of the E2E-ABSA task, if the same aspect term appears with different polarities in one review, we assign it the \textit{neutral} polarity. Prediction for both tasks is considered correct only if the boundary (and sentiment polarity in the case of E2E-ABSA) are correct.

\par Three Czech models (Czert, RobeCzech and FERNET) are further pre-trained using the masked language modelling (MLM) task \citep{devlin-etal-2019-bert} with the intention to adapt them to the task domain and improve the overall results.

\subsection{Sequence-to-Sequence Models}
We employ sequence-to-sequence models to perform ACD, ATE, ACTE and TASD tasks simultaneously. These models process text as input and produce text as output. To the best of our knowledge, no monolingual sequence-to-sequence models are currently explicitly designed for Czech. Consequently, we have decided to use the large \textbf{mT5} model \citep{xue-etal-2021-mt5} and the large \textbf{mBART} model \citep{tang-etal-2021-multilingual}. These models are the multilingual adaptations of the English T5 \citep{raffel2020exploring} and BART \citep{lewis-etal-2020-bart} models.
\par
The sequence-to-sequence models consist of two parts of the Transformer architecture: the \textit{encoder} and the \textit{decoder}. The encoder transforms input sequence $x$ into a contextualized sequence $\vec{e}$. Given the encoded input $\vec{e}$, the decoder models the conditional probability distribution $P_{\vec{\Theta}}(y|\vec{e})$ of the target sequence $y$, where $\vec{\Theta}$ are the model's parameters. The decoder calculates the output $y_i$ at each step $i$ based on the previous outputs $y_1, \ldots, y_{i-1}$ and the encoded input $\vec{e}$.
\par
Since the output of sequence-to-sequence models is text, we convert discrete ABSA labels to textual format, \textcolor{black}{following \citet{smid-priban-2023-prompt}}. The label is constructed as \quotes{\textit{$c$ is $P_p(p)$, given the expression: $a$}}, where $a$ is the aspect term, $c$ is the aspect category, and $P_p(p)$ is a mapping function that maps the sentiment polarity $p$ as
\begin{equation}
    \small
    P_p(p) = 
    \begin{cases}
        \textit{great} & \text{if $p$ is \textit{positive},}\\
        \textit{ok} & \text{if $p$ is \textit{neutral},}\\
        \textit{bad} & \text{if $p$ is \textit{negative}.}\\
    \end{cases}
\end{equation}
For example, the review \textit{\quotes{Výtečné pivo} (\quotes{Excellent beer})} yields the label \quotes{\textit{Drinks quality} is \textit{great}, given the expression: \textit{pivo}}. Multiple sentiment triplets in reviews are concatenated with semicolons.
\par
In this context, the model takes the text (review) as input and aims to generate a textual label as its output. The model's parameters are fine-tuned to optimize label generation in the specified format. The model always generates all outputs together, i.e. for the TASD task, from which specific elements required for other tasks are extracted, e.g. category for the ACD task. We discard \quotes{NULL} targets for the ATE task and ignore duplicate targets for the ATE and ACD tasks, as in \citet{pontiki-etal-2016-semeval}.
\par
Since our approach predicts the aspect category and term alongside sentiment, we do not use these models for the APD task, which assumes the model has access to the gold data for aspect terms and categories. A fair comparison would require modifications of the input and output formats.

Similarly, we refrain from using these models for the E2E-ABSA task as their results cannot be fairly compared with encoder-based models. Sequence-to-sequence models can predict \quotes{NULL} terms and generate one aspect term multiple times with different polarities. In contrast, encoder-based models predict only one polarity for a single aspect term and do not predict the \quotes{NULL} aspect term.

\subsection{Hyperparameters}
We train the models with various hyperparameters. We use a batch size of 64 for each experiment and search for the learning rate from \{3e-4, 1e-4, 5e-5, 1e-5\}. Encoder-based models run for up to 50 epochs, while sequence-to-sequence models run for up to 35 epochs, using the greedy encoding algorithm for simplicity. We employ the AdamW optimizer \citep{loshchilov2017decoupled} for all the models except the mT5 model, where we use the Adafactor optimizer \citep{shazeer2018adafactor}.
\par
We then select the best-performing models on validation data, fine-tune them on merged training and validation data and evaluate them on the test data. We conduct each experiment five times, each with a different random seed, to ensure the reliability of our results. We present the average scores along with a 95\% confidence interval.

\par We also use the AdamW optimizer and the cross-entropy loss function for the additional MLM pre-training. The word masking probability is set to 15\%. We pre-train the model for 20K steps with a~batch size of 512 and a maximum input length set to 512 tokens with a learning rate of 5e-5.

\section{Results}
Table \ref{tab:res} shows the results achieved by the encoder-based models. We can see that the multilingual XLM-R models (particularly the large version, which has the most parameters out of all these models) perform similarly to Czech-only (monolingual) models. In some cases, they outperform them. The easiest task is the APD task, where the models assign only one of three classes. The ACD task is more complex than the APD task because the models have to choose from more classes, and the problem is multi-label. E2E-ABSA is the most challenging task because the model has to assign the correct class to each token and correctly predict the aspect term boundaries alongside the sentiment polarity. The ATE task is less difficult than the E2E-ABSA task because the model does not have to assign the sentiment polarity for the tokens. These claims are supported by the reported results corresponding to the different difficulty levels of each task; the easiest tasks achieve much better results than the more difficult ones. The baseline results shown in Table \ref{tab:res} achieved by \citet{hercig2016unsupervised} are on the old dataset, which has different annotations and aspect categories.
\par
Overall, the results for the \texttt{CsRest-O} dataset are generally worse than for the two remaining datasets, possibly due to the smaller training data size. While there are some differences between the results for \texttt{CsRest-N} and \texttt{CsRest-M} datasets for each task and model, it is unclear whether either version is consistently more challenging.

\begin{table*}[ht!]
    \centering
    \begin{adjustbox}{width=1\linewidth, center}
         \begin{tabular}{lrrrrrrrrrrrr}
            \toprule
            \multirow{2}{*}{\textbf{Model}}             & \multicolumn{4}{c}{\textbf{\texttt{CsRest-O}}}                                      & \multicolumn{4}{c}{\textbf{\texttt{CsRest-N}}}                                                                                                                                                          & \multicolumn{4}{c}{\textbf{\texttt{CsRest-M}}}                                                                                                                                                        \\ \cmidrule(lr){2-5} \cmidrule(lr){6-9} \cmidrule(lr){10-13}
                                            & APD             & ACD             & ATE             & E2E             & APD                                          & ACD                                          & ATE                                          & E2E                                          & APD                                          & ACD                                          & ATE                                          & E2E                                          \\ \midrule
            Czert                                       & 83.2$^{\pm1.4}$ & 81.2$^{\pm1.4}$ & 81.7$^{\pm0.4}$ & 66.8$^{\pm0.7}$ & 85.5$^{\pm4.9}$                              & 81.6$^{\pm1.5}$                              & 78.4$^{\pm1.0}$                              & 70.9$^{\pm1.2}$                              & 85.3$^{\pm0.9}$                              & 82.2$^{\pm0.3}$                              & 82.8$^{\pm0.7}$                              & 70.6$^{\pm0.9}$                              \\
            RobeCzech                                   & 85.2$^{\pm1.6}$ & 80.9$^{\pm2.5}$ & 82.9$^{\pm0.4}$ & 67.8$^{\pm1.6}$ & 89.4$^{\pm1.2}$                              & 80.8$^{\pm1.6}$                              & 78.8$^{\pm1.1}$                              & 71.9$^{\pm1.6}$                              & 87.6$^{\pm1.3}$                              & 83.1$^{\pm1.0}$                              & 82.8$^{\pm0.5}$                              & 71.3$^{\pm1.9}$                              \\
            FERNET                                      & \underline{86.0}$^{\pm0.4}$ & \underline{83.7}$^{\pm1.2}$ & \textbf{84.9}$^{\pm1.1}$ & \underline{71.7}$^{\pm2.1}$ & \underline{90.1}$^{\pm2.2}$ & \textbf{82.9}$^{\pm0.9}$    & \textbf{80.8}$^{\pm1.2}$    & \underline{74.7}$^{\pm1.5}$ & \textbf{88.2}$^{\pm0.8}$    & \underline{84.3}$^{\pm0.4}$ & \underline{83.2}$^{\pm1.1}$ & \textbf{74.8}$^{\pm1.1}$    \\
            Small-E-Czech                               & 78.0$^{\pm3.7}$ & 75.5$^{\pm1.2}$ & 81.5$^{\pm0.9}$ & 59.2$^{\pm4.8}$ & 84.6$^{\pm1.6}$                              & 76.7$^{\pm2.0}$                              & 77.3$^{\pm2.4}$                              & 64.0$^{\pm2.9}$                              & 83.3$^{\pm0.8}$                              & 79.8$^{\pm0.7}$                              & 81.1$^{\pm0.7}$                              & 66.7$^{\pm2.0}$                              \\
            mBERT                                       & 77.1$^{\pm3.8}$ & 77.8$^{\pm2.1}$ & 79.6$^{\pm0.6}$ & 60.3$^{\pm2.8}$ & 85.1$^{\pm1.8}$                              & 78.6$^{\pm1.3}$                              & 76.2$^{\pm1.7}$                              & 67.5$^{\pm2.0}$                              & 82.2$^{\pm1.0}$                              & 79.0$^{\pm0.5}$                              & 80.0$^{\pm0.6}$                              & 67.7$^{\pm1.6}$                              \\
            XLM-R\textsubscript{BASE}  & 80.7$^{\pm2.3}$ & 80.4$^{\pm1.4}$ & 82.4$^{\pm0.6}$ & 68.9$^{\pm3.6}$ & 88.5$^{\pm1.8}$                              & 80.6$^{\pm1.7}$                              & 78.6$^{\pm1.3}$                              & 70.7$^{\pm2.1}$                              & 85.1$^{\pm1.6}$                              & 81.0$^{\pm1.1}$                              & 82.0$^{\pm0.4}$                              & 70.4$^{\pm0.7}$                              \\
            XLM-R\textsubscript{LARGE} & \textbf{87.2}$^{\pm1.5}$ & \textbf{85.7}$^{\pm0.4}$ & \underline{84.0}$^{\pm0.8}$ & \textbf{71.9}$^{\pm2.3}$ & \textbf{91.4}$^{\pm0.9}$    & \underline{82.8}$^{\pm1.0}$ & \underline{80.2}$^{\pm1.1}$ & \textbf{75.5}$^{\pm1.0}$    & \underline{87.9}$^{\pm0.8}$ & \textbf{86.2}$^{\pm0.3}$    & \textbf{83.5}$^{\pm1.2}$    & \underline{74.4}$^{\pm1.0}$ \\ \cdashlinelr{1-13}
            Czert*                                       & 88.4$^{\pm0.7}$ & 86.8$^{\pm0.9}$ & 85.7$^{\pm1.7}$ & 74.7$^{\pm1.4}$ & 89.2$^{\pm2.6}$                              & 84.6$^{\pm0.4}$                              & 81.3$^{\pm1.4}$                              & 73.8$^{\pm1.2}$                              & 88.3$^{\pm1.1}$                              & 86.1$^{\pm0.5}$                              & 84.4$^{\pm1.0}$                              & 75.6$^{\pm0.5}$                              \\
            RobeCzech*                                   & 88.4$^{\pm0.9}$ & 84.9$^{\pm0.7}$ & 85.3$^{\pm1.1}$ & 70.4$^{\pm2.3}$ & 91.1$^{\pm0.8}$                              & 83.9$^{\pm0.7}$                              & 82.3$^{\pm1.0}$                              & 74.3$^{\pm0.7}$                              & 88.4$^{\pm0.8}$                              & 85.7$^{\pm0.9}$                              & 84.9$^{\pm1.2}$                              & 75.4$^{\pm1.1}$                              \\
            FERNET*  & 85.0$^{\pm1.1}$ & 83.9$^{\pm0.7}$ & 84.0$^{\pm0.9}$ & 71.7$^{\pm1.0}$ & 91.0$^{\pm1.5}$                              & 84.0$^{\pm1.5}$                              & 82.3$^{\pm1.2}$                              & 75.9$^{\pm0.8}$                              & 90.0$^{\pm0.5}$                              & 87.1$^{\pm0.4}$                              & 85.6$^{\pm0.8}$                              & 77.0$^{\pm0.4}$                              \\     
            \cdashlinelr{1-13}
            & \multicolumn{1}{c}{-} & 80.0\phantom{$^{\pm0.0}$} & 78.7\phantom{$^{\pm0.0}$}  & \multicolumn{1}{c}{-} & \multicolumn{1}{c}{-} & \multicolumn{1}{c}{-} & \multicolumn{1}{c}{-} & \multicolumn{1}{c}{-} & \multicolumn{1}{c}{-} & \multicolumn{1}{c}{-} & \multicolumn{1}{c}{-} & \multicolumn{1}{c}{-} \\
            \bottomrule
        \end{tabular}
    \end{adjustbox}
    \caption{F1 scores for the new Czech ABSA dataset. The best score for each task and dataset version is in \textbf{bold}; the second best is \underline{underlined}. Models marked with * are additionally pre-trained on the unsupervised dataset and are not considered for the best results. The $^\dagger$ symbol denotes results by \citet{hercig2016unsupervised} obtained for the old dataset with different annotations and aspect categories.}
    \label{tab:res}
\end{table*}

\par Additionally, we pre-trained three Czech models on the unsupervised dataset. The results show that the additional pre-training significantly improves the performance of all three models.

For example, the RobeCzech model shows an improvement of approximately 4\% on the E2E-ABSA task and \texttt{CsRest-M} dataset.

\par

\par
Table \ref{tab:res_seq2seq} shows the results of sequence-to-sequence models. The mBART model outperforms the mT5 model on all tasks. The mT5 model also performs the best on the \texttt{CsRest-O} dataset compared to the other versions. The mBART model performs similarly on all versions. Worse results of the mT5 model could imply that a better hyperparameters search should be done for the mT5 model. Overall, the TASD task is the most challenging because the model must simultaneously predict the aspect term, aspect category and sentiment polarity correctly.
\begin{table}[ht!]
    \centering
    \begin{adjustbox}{width=0.95\linewidth, center}
        \begin{tabular}{@{}llrrr@{}}
            \toprule
            \textbf{Model}         & \textbf{Task} & \textbf{\texttt{CsRest-O}} & \textbf{\texttt{CsRest-N}}      & \textbf{\texttt{CsRest-M}}              \\ \midrule
            \multirow{4}{*}{mT5}   & ACD           & 75.4$^{\pm1.8}$ & 68.9$^{\pm1.1}$ & 70.8$^{\pm1.5}$        \\
                                   & ATE           & 66.5$^{\pm2.5}$ & 59.7$^{\pm1.5}$ & 66.9$^{\pm1.4}$        \\
                                   & ACTE          & 56.4$^{\pm1.0}$ & 45.0$^{\pm1.7}$ & 52.6$^{\pm1.8}$        \\
                                   & TASD          & 48.0$^{\pm1.0}$ & 41.1$^{\pm1.8}$ & 46.4$^{\pm1.5}$        \\ \cdashlinelr{1-5}
            \multirow{4}{*}{mBART} & ACD           & \textbf{78.7}$^{\pm1.6}$ & \textbf{79.3}$^{\pm0.4}$ & \textbf{80.6}$^{\pm1.7}$        \\
                                   & ATE           & \textbf{78.9}$^{\pm1.3}$ & \textbf{76.0}$^{\pm1.5}$ & \textbf{79.7}$^{\pm1.1}$        \\
                                   & ACTE          & \textbf{67.2}$^{\pm1.4}$ & \textbf{62.4}$^{\pm0.7}$ & \textbf{67.3}$^{\pm1.2}$        \\
                                   & TASD          & \textbf{57.5}$^{\pm1.7}$ & \textbf{56.3}$^{\pm0.6}$ & \textbf{59.3}$^{\pm1.4}$        \\ \bottomrule
        \end{tabular}
    \end{adjustbox}
    \caption{F1 scores for different models across tasks on the new Czech ABSA dataset. The best result for each task and dataset version is in \textbf{bold}.}
    \label{tab:res_seq2seq}
\end{table}

\par
The encoder-based models consistently outperform the sequence-to-sequence models.
The reason may be that the encoder-based models are always specialized directly for one task. On the other hand, the sequence-to-sequence models generate the output for the TASD task. Then, we extract only the relevant elements for the specific task from the output (e.g. only aspect terms for the ATE task).

\subsection{Error Analysis}
We conducted an error analysis of the sequence-to-sequence models to understand the key characteristics of our dataset and identify the main challenges these models face. Our findings revealed several important observations:
\par
\textbf{Output format:} The mT5 model occasionally struggles to produce the correct output format, which is crucial for target extraction. On the other hand, the mBART model makes this error to a lesser extent, possibly contributing to its superior performance over mT5. Both models frequently generate duplicate outputs, reducing the diversity of generated sentiment triplets. While we ensure not to count identical triplets multiple times (thus not impacting the results), this repetition restricts the models from generating unique outputs, potentially causing them to miss specific prediction targets.
\par
\textbf{Aspect term prediction:} Both models encounter challenges in predicting the correct aspect terms. They sometimes generate only a part of the aspect term rather than the complete term (e.g. \textit{\quotes{burrito}} instead of \textit{\quotes{burrito bowl}}). Additionally, the models may blend parts of the review, leading to outputs that do not match the original text's specific form. For example, instead of \textit{\quotes{raspberries with ice cream and whipped cream}}, the model generates \textit{\quotes{raspberries with whipped cream}}, a phrase not present in the original review.
\par
\textbf{Handling typos:}
The models generate words in the correct form even when they appear as typos in the original review. For instance, if the review contains the typo \textit{\quotes{se\textbf{vr}ice}}, the model generates the corrected word \textit{\quotes{service}}.
The models also occasionally produce lowercase output even when the original text contains uppercase letters.
\par
\textbf{Making up words:} The models sometimes make up words not found in the reviews. For example, some reviews imply opinions about the ambience, and the models may generate \textit{\quotes{ambience}} instead of \quotes{NULL} as an aspect term.
\par
\textbf{Aspect category confusion:} Regarding aspect categories, the models frequently omit the less common categories, such as \textit{LOCATION\#GENERAL} or \textit{DRINKS\#STYLE\_OPTIONS}. Both models often confuse the \textit{RESTAURANT\#MISCELLANEOUS} and \textit{RESTAURANT\#GENERAL} classes.
\par
\textbf{Sentiment polarity challenges:} The most significant challenge arises with neutral sentiment polarity. Despite being the least frequent class, both models rarely predict it and tend to predict either negative or positive sentiment.

\section{Conclusion}
In this paper, we present a novel manually annotated Czech dataset in the restaurant domain for aspect-based sentiment analysis. The dataset comes in three different versions and is the largest of its kind in the Czech language. Unlike the previous Czech ABSA datasets, this newly created dataset establishes connections between multiple sentiment elements, allowing for solving more complex ABSA tasks, such as the TASD task. Notably, our dataset adheres to the same format as the SemEval-2016 dataset, potentially enabling cross-lingual experiments in the future. Next, we provide large unlabelled corpora 
for unsupervised training.

\par We also provide strong baseline results for various ABSA tasks utilizing models based on the Transformer architecture. Our system is language and domain-independent, meaning it can easily be trained on data from other languages. Our research extends beyond the numerical outcomes, delving into an insightful error analysis that elucidates the unique challenges and limitations our dataset poses to these models.

\par In summary, our study not only provides a new ABSA dataset for the Czech language but also establishes a benchmark for Czech ABSA research. We anticipate that this resource will catalyze future research endeavours, advancing the field of sentiment analysis and fostering cross-lingual exploration within the ABSA domain. 

\section*{Acknowledgements}
This work has been partly supported by grant No. SGS-2022-016 Advanced methods of data processing and analysis.
Computational resources were provided by the e-INFRA CZ project (ID:90254), supported by the Ministry of Education, Youth and Sports of the Czech Republic.

\section{Bibliographical References}\label{sec:reference}

\bibliographystyle{lrec-coling2024-natbib}
\bibliography{lrec-coling2024}

\begin{thebibliography}{36}
\expandafter\ifx\csname natexlab\endcsname\relax\def\natexlab#1{#1}\fi

\bibitem[{Conneau et~al.(2020)Conneau, Khandelwal, Goyal, Chaudhary, Wenzek, Guzm{\'a}n, Grave, Ott, Zettlemoyer, and Stoyanov}]{conneau-etal-2020-unsupervised}
Alexis Conneau, Kartikay Khandelwal, Naman Goyal, Vishrav Chaudhary, Guillaume Wenzek, Francisco Guzm{\'a}n, Edouard Grave, Myle Ott, Luke Zettlemoyer, and Veselin Stoyanov. 2020.
\newblock \href {https://doi.org/10.18653/v1/2020.acl-main.747} {Unsupervised cross-lingual representation learning at scale}.
\newblock In \emph{Proceedings of the 58th Annual Meeting of the Association for Computational Linguistics}, pages 8440--8451, Online. Association for Computational Linguistics.

\bibitem[{Devlin et~al.(2019)Devlin, Chang, Lee, and Toutanova}]{devlin-etal-2019-bert}
Jacob Devlin, Ming-Wei Chang, Kenton Lee, and Kristina Toutanova. 2019.
\newblock \href {https://doi.org/10.18653/v1/N19-1423} {{BERT}: Pre-training of deep bidirectional transformers for language understanding}.
\newblock In \emph{Proceedings of the 2019 Conference of the North {A}merican Chapter of the Association for Computational Linguistics: Human Language Technologies, Volume 1 (Long and Short Papers)}, pages 4171--4186, Minneapolis, Minnesota. Association for Computational Linguistics.

\bibitem[{Dong et~al.(2014)Dong, Wei, Tan, Tang, Zhou, and Xu}]{dong-etal-2014-adaptive}
Li~Dong, Furu Wei, Chuanqi Tan, Duyu Tang, Ming Zhou, and Ke~Xu. 2014.
\newblock \href {https://doi.org/10.3115/v1/P14-2009} {Adaptive recursive neural network for target-dependent {T}witter sentiment classification}.
\newblock In \emph{Proceedings of the 52nd Annual Meeting of the Association for Computational Linguistics (Volume 2: Short Papers)}, pages 49--54, Baltimore, Maryland. Association for Computational Linguistics.

\bibitem[{Fan et~al.(2019)Fan, Wu, Dai, Huang, and Chen}]{fan-etal-2019-target}
Zhifang Fan, Zhen Wu, Xin-Yu Dai, Shujian Huang, and Jiajun Chen. 2019.
\newblock \href {https://doi.org/10.18653/v1/N19-1259} {Target-oriented opinion words extraction with target-fused neural sequence labeling}.
\newblock In \emph{Proceedings of the 2019 Conference of the North {A}merican Chapter of the Association for Computational Linguistics: Human Language Technologies, Volume 1 (Long and Short Papers)}, pages 2509--2518, Minneapolis, Minnesota. Association for Computational Linguistics.

\bibitem[{Gou et~al.(2023)Gou, Guo, and Yang}]{gou-etal-2023-mvp}
Zhibin Gou, Qingyan Guo, and Yujiu Yang. 2023.
\newblock \href {https://doi.org/10.18653/v1/2023.acl-long.240} {{M}v{P}: Multi-view prompting improves aspect sentiment tuple prediction}.
\newblock In \emph{Proceedings of the 61st Annual Meeting of the Association for Computational Linguistics (Volume 1: Long Papers)}, pages 4380--4397, Toronto, Canada. Association for Computational Linguistics.

\bibitem[{Hercig et~al.(2016)Hercig, Brychc{\'\i}n, Svoboda, Konkol, and Steinberger}]{hercig2016unsupervised}
Tom{\'a}{\v{s}} Hercig, Tom{\'a}{\v{s}} Brychc{\'\i}n, Luk{\'a}{\v{s}} Svoboda, Michal Konkol, and Josef Steinberger. 2016.
\newblock Unsupervised methods to improve aspect-based sentiment analysis in czech.
\newblock \emph{Computaci{\'o}n y Sistemas}, 20(3):365--375.

\bibitem[{Hyun et~al.(2020)Hyun, Cho, and Yu}]{hyun-etal-2020-building}
Dongmin Hyun, Junsu Cho, and Hwanjo Yu. 2020.
\newblock \href {https://doi.org/10.18653/v1/2020.coling-main.83} {Building large-scale {E}nglish and {K}orean datasets for aspect-level sentiment analysis in automotive domain}.
\newblock In \emph{Proceedings of the 28th International Conference on Computational Linguistics}, pages 961--966, Barcelona, Spain (Online). International Committee on Computational Linguistics.

\bibitem[{Jiang et~al.(2019)Jiang, Chen, Xu, Ao, and Yang}]{jiang-etal-2019-challenge}
Qingnan Jiang, Lei Chen, Ruifeng Xu, Xiang Ao, and Min Yang. 2019.
\newblock \href {https://doi.org/10.18653/v1/D19-1654} {A challenge dataset and effective models for aspect-based sentiment analysis}.
\newblock In \emph{Proceedings of the 2019 Conference on Empirical Methods in Natural Language Processing and the 9th International Joint Conference on Natural Language Processing (EMNLP-IJCNLP)}, pages 6280--6285, Hong Kong, China. Association for Computational Linguistics.

\bibitem[{Koci{\'{a}}n et~al.(2022)Koci{\'{a}}n, N{\'{a}}plava, Stancl, and Kadlec}]{kocian2022siamese}
Matej Koci{\'{a}}n, Jakub N{\'{a}}plava, Daniel Stancl, and Vladim{\'{\i}}r Kadlec. 2022.
\newblock \href {https://doi.org/10.1609/aaai.v36i11.21502} {Siamese bert-based model for web search relevance ranking evaluated on a new czech dataset}.
\newblock In \emph{Thirty-Sixth {AAAI} Conference on Artificial Intelligence, {AAAI} 2022, Thirty-Fourth Conference on Innovative Applications of Artificial Intelligence, {IAAI} 2022, The Twelveth Symposium on Educational Advances in Artificial Intelligence, {EAAI} 2022 Virtual Event, February 22 - March 1, 2022}, pages 12369--12377. {AAAI} Press.

\bibitem[{Lehe{\v{c}}ka and {\v{S}}vec(2021)}]{FERNET-2021}
Jan Lehe{\v{c}}ka and Jan {\v{S}}vec. 2021.
\newblock Comparison of czech transformers on text classification tasks.
\newblock In \emph{Statistical Language and Speech Processing}, pages 27--37, Cham. Springer International Publishing.

\bibitem[{Lenc and Hercig(2016)}]{SLON-Lenc2016Neural}
Ladislav Lenc and Tom{\'a}s Hercig. 2016.
\newblock Neural networks for sentiment analysis in czech.
\newblock In \emph{Proceedings of the 16th {ITAT}: Slovensko{\v{c}}esk{\'{y}} {NLP} workshop (Slo{NLP} 2016)}, volume 1649 of \emph{{CEUR} Workshop Proceedings}, pages 48--55, Bratislava, Slovakia. Comenius University in Bratislava, Faculty of Mathematics, Physics and Informatics, CreateSpace Independent Publishing Platform.

\bibitem[{Lewis et~al.(2020)Lewis, Liu, Goyal, Ghazvininejad, Mohamed, Levy, Stoyanov, and Zettlemoyer}]{lewis-etal-2020-bart}
Mike Lewis, Yinhan Liu, Naman Goyal, Marjan Ghazvininejad, Abdelrahman Mohamed, Omer Levy, Veselin Stoyanov, and Luke Zettlemoyer. 2020.
\newblock \href {https://doi.org/10.18653/v1/2020.acl-main.703} {{BART}: Denoising sequence-to-sequence pre-training for natural language generation, translation, and comprehension}.
\newblock In \emph{Proceedings of the 58th Annual Meeting of the Association for Computational Linguistics}, pages 7871--7880, Online. Association for Computational Linguistics.

\bibitem[{Li et~al.(2019)Li, Bing, Zhang, and Lam}]{li-etal-2019-exploiting}
Xin Li, Lidong Bing, Wenxuan Zhang, and Wai Lam. 2019.
\newblock \href {https://doi.org/10.18653/v1/D19-5505} {Exploiting {BERT} for end-to-end aspect-based sentiment analysis}.
\newblock In \emph{Proceedings of the 5th Workshop on Noisy User-generated Text (W-NUT 2019)}, pages 34--41, Hong Kong, China. Association for Computational Linguistics.

\bibitem[{Liu(2012)}]{liu2012sentiment}
Bing Liu. 2012.
\newblock Sentiment analysis and opinion mining.
\newblock \emph{Synthesis lectures on human language technologies}, 5(1):1--167.

\bibitem[{Loshchilov and Hutter(2017)}]{loshchilov2017decoupled}
Ilya Loshchilov and Frank Hutter. 2017.
\newblock Decoupled weight decay regularization.
\newblock \emph{arXiv preprint arXiv:1711.05101}.

\bibitem[{Nakayama et~al.(2022)Nakayama, Murakami, Kumar, Bhingardive, and Hardaway}]{nakayama-etal-2022-large}
Yuki Nakayama, Koji Murakami, Gautam Kumar, Sudha Bhingardive, and Ikuko Hardaway. 2022.
\newblock \href {https://aclanthology.org/2022.lrec-1.758} {A large-scale {J}apanese dataset for aspect-based sentiment analysis}.
\newblock In \emph{Proceedings of the Thirteenth Language Resources and Evaluation Conference}, pages 7014--7021, Marseille, France. European Language Resources Association.

\bibitem[{Pontiki et~al.(2016)Pontiki, Galanis, Papageorgiou, Androutsopoulos, Manandhar, AL-Smadi, Al-Ayyoub, Zhao, Qin, De~Clercq, Hoste, Apidianaki, Tannier, Loukachevitch, Kotelnikov, Bel, Jim{\'e}nez-Zafra, and Eryi{\u{g}}it}]{pontiki-etal-2016-semeval}
Maria Pontiki, Dimitris Galanis, Haris Papageorgiou, Ion Androutsopoulos, Suresh Manandhar, Mohammad AL-Smadi, Mahmoud Al-Ayyoub, Yanyan Zhao, Bing Qin, Orph{\'e}e De~Clercq, V{\'e}ronique Hoste, Marianna Apidianaki, Xavier Tannier, Natalia Loukachevitch, Evgeniy Kotelnikov, Nuria Bel, Salud~Mar{\'\i}a Jim{\'e}nez-Zafra, and G{\"u}l{\c{s}}en Eryi{\u{g}}it. 2016.
\newblock \href {https://doi.org/10.18653/v1/S16-1002} {{S}em{E}val-2016 task 5: Aspect based sentiment analysis}.
\newblock In \emph{Proceedings of the 10th International Workshop on Semantic Evaluation ({S}em{E}val-2016)}, pages 19--30, San Diego, California. Association for Computational Linguistics.

\bibitem[{Pontiki et~al.(2015)Pontiki, Galanis, Papageorgiou, Manandhar, and Androutsopoulos}]{pontiki-etal-2015-semeval}
Maria Pontiki, Dimitris Galanis, Haris Papageorgiou, Suresh Manandhar, and Ion Androutsopoulos. 2015.
\newblock \href {https://doi.org/10.18653/v1/S15-2082} {{S}em{E}val-2015 task 12: Aspect based sentiment analysis}.
\newblock In \emph{Proceedings of the 9th International Workshop on Semantic Evaluation ({S}em{E}val 2015)}, pages 486--495, Denver, Colorado. Association for Computational Linguistics.

\bibitem[{Pontiki et~al.(2014)Pontiki, Galanis, Pavlopoulos, Papageorgiou, Androutsopoulos, and Manandhar}]{pontiki-etal-2014-semeval}
Maria Pontiki, Dimitris Galanis, John Pavlopoulos, Harris Papageorgiou, Ion Androutsopoulos, and Suresh Manandhar. 2014.
\newblock \href {https://doi.org/10.3115/v1/S14-2004} {{S}em{E}val-2014 task 4: Aspect based sentiment analysis}.
\newblock In \emph{Proceedings of the 8th International Workshop on Semantic Evaluation ({S}em{E}val 2014)}, pages 27--35, Dublin, Ireland. Association for Computational Linguistics.

\bibitem[{P{\v{r}}ib{\'a}{\v{n}} and Pra\v{z}{\'a}k(2023)}]{priban-prazak-2023-improving}
Pavel P{\v{r}}ib{\'a}{\v{n}} and Ond\v{r}ej Pra\v{z}{\'a}k. 2023.
\newblock \href {https://aclanthology.org/2023.ranlp-1.96} {Improving aspect-based sentiment with end-to-end semantic role labeling model}.
\newblock In \emph{Proceedings of the 14th International Conference on Recent Advances in Natural Language Processing}, pages 888--897, Varna, Bulgaria. INCOMA Ltd., Shoumen, Bulgaria.

\bibitem[{Raffel et~al.(2020)Raffel, Shazeer, Roberts, Lee, Narang, Matena, Zhou, Li, and Liu}]{raffel2020exploring}
Colin Raffel, Noam Shazeer, Adam Roberts, Katherine Lee, Sharan Narang, Michael Matena, Yanqi Zhou, Wei Li, and Peter~J Liu. 2020.
\newblock Exploring the limits of transfer learning with a unified text-to-text transformer.
\newblock \emph{The Journal of Machine Learning Research}, 21(1):5485--5551.

\bibitem[{Saeidi et~al.(2016)Saeidi, Bouchard, Liakata, and Riedel}]{saeidi-etal-2016-sentihood}
Marzieh Saeidi, Guillaume Bouchard, Maria Liakata, and Sebastian Riedel. 2016.
\newblock \href {https://aclanthology.org/C16-1146} {{S}enti{H}ood: Targeted aspect based sentiment analysis dataset for urban neighbourhoods}.
\newblock In \emph{Proceedings of {COLING} 2016, the 26th International Conference on Computational Linguistics: Technical Papers}, pages 1546--1556, Osaka, Japan. The COLING 2016 Organizing Committee.

\bibitem[{Shazeer and Stern(2018)}]{shazeer2018adafactor}
Noam Shazeer and Mitchell Stern. 2018.
\newblock \href {http://arxiv.org/abs/1804.04235} {Adafactor: Adaptive learning rates with sublinear memory cost}.

\bibitem[{Sido et~al.(2021)Sido, Pra{\v{z}}{\'a}k, P{\v{r}}ib{\'a}{\v{n}}, Pa{\v{s}}ek, Sej{\'a}k, and Konop{\'\i}k}]{sido2021czert}
Jakub Sido, Ond{\v{r}}ej Pra{\v{z}}{\'a}k, Pavel P{\v{r}}ib{\'a}{\v{n}}, Jan Pa{\v{s}}ek, Michal Sej{\'a}k, and Miloslav Konop{\'\i}k. 2021.
\newblock \href {https://aclanthology.org/2021.ranlp-1.149} {Czert {--} {C}zech {BERT}-like model for language representation}.
\newblock In \emph{Proceedings of the International Conference on Recent Advances in Natural Language Processing (RANLP 2021)}, pages 1326--1338, Held Online. INCOMA Ltd.

\bibitem[{{\v{S}}m{\'\i}d and P{\v{r}}ib{\'a}{\v{n}}(2023)}]{smid-priban-2023-prompt}
Jakub {\v{S}}m{\'\i}d and Pavel P{\v{r}}ib{\'a}{\v{n}}. 2023.
\newblock \href {https://aclanthology.org/2023.ranlp-1.118} {Prompt-based approach for {C}zech sentiment analysis}.
\newblock In \emph{Proceedings of the 14th International Conference on Recent Advances in Natural Language Processing}, pages 1110--1120, Varna, Bulgaria. INCOMA Ltd., Shoumen, Bulgaria.

\bibitem[{Steinberger et~al.(2014)Steinberger, Brychc{\'\i}n, and Konkol}]{steinberger-etal-2014-aspect}
Josef Steinberger, Tom{\'a}{\v{s}} Brychc{\'\i}n, and Michal Konkol. 2014.
\newblock \href {https://doi.org/10.3115/v1/W14-2605} {Aspect-level sentiment analysis in {C}zech}.
\newblock In \emph{Proceedings of the 5th Workshop on Computational Approaches to Subjectivity, Sentiment and Social Media Analysis}, pages 24--30, Baltimore, Maryland. Association for Computational Linguistics.

\bibitem[{Straka et~al.(2021)Straka, N{\'a}plava, Strakov{\'a}, and Samuel}]{straka2021robeczech}
Milan Straka, Jakub N{\'a}plava, Jana Strakov{\'a}, and David Samuel. 2021.
\newblock {RobeCzech: Czech RoBERTa, a Monolingual Contextualized Language Representation Model}.
\newblock In \emph{Text, Speech, and Dialogue}, pages 197--209, Cham. Springer International Publishing.

\bibitem[{Tamchyna et~al.(2015)Tamchyna, Fiala, and Veselovsk{\'a}}]{tamchyna2015czech}
Ales Tamchyna, Ondrej Fiala, and Katerina Veselovsk{\'a}. 2015.
\newblock Czech aspect-based sentiment analysis: A new dataset and preliminary results.
\newblock In \emph{ITAT}, pages 95--99.

\bibitem[{Tang et~al.(2021)Tang, Tran, Li, Chen, Goyal, Chaudhary, Gu, and Fan}]{tang-etal-2021-multilingual}
Yuqing Tang, Chau Tran, Xian Li, Peng-Jen Chen, Naman Goyal, Vishrav Chaudhary, Jiatao Gu, and Angela Fan. 2021.
\newblock \href {https://doi.org/10.18653/v1/2021.findings-acl.304} {Multilingual translation from denoising pre-training}.
\newblock In \emph{Findings of the Association for Computational Linguistics: ACL-IJCNLP 2021}, pages 3450--3466, Online. Association for Computational Linguistics.

\bibitem[{Vaswani et~al.(2017)Vaswani, Shazeer, Parmar, Uszkoreit, Jones, Gomez, Kaiser, and Polosukhin}]{vaswani2017attention}
Ashish Vaswani, Noam Shazeer, Niki Parmar, Jakob Uszkoreit, Llion Jones, Aidan~N Gomez, {\L}ukasz Kaiser, and Illia Polosukhin. 2017.
\newblock Attention is all you need.
\newblock \emph{Advances in neural information processing systems}, 30.

\bibitem[{Wan et~al.(2020)Wan, Yang, Du, Liu, Qi, and Pan}]{tasd}
Hai Wan, Yufei Yang, Jianfeng Du, Yanan Liu, Kunxun Qi, and Jeff~Z. Pan. 2020.
\newblock \href {https://doi.org/10.1609/aaai.v34i05.6447} {Target-aspect-sentiment joint detection for aspect-based sentiment analysis}.
\newblock \emph{Proceedings of the AAAI Conference on Artificial Intelligence}, 34(05):9122--9129.

\bibitem[{Wang et~al.(2018)Wang, Lan, and Wang}]{wang2018towards}
Feixiang Wang, Man Lan, and Wenting Wang. 2018.
\newblock Towards a one-stop solution to both aspect extraction and sentiment analysis tasks with neural multi-task learning.
\newblock In \emph{2018 International joint conference on neural networks (IJCNN)}, pages 1--8. IEEE.

\bibitem[{Xue et~al.(2021)Xue, Constant, Roberts, Kale, Al-Rfou, Siddhant, Barua, and Raffel}]{xue-etal-2021-mt5}
Linting Xue, Noah Constant, Adam Roberts, Mihir Kale, Rami Al-Rfou, Aditya Siddhant, Aditya Barua, and Colin Raffel. 2021.
\newblock \href {https://doi.org/10.18653/v1/2021.naacl-main.41} {m{T}5: A massively multilingual pre-trained text-to-text transformer}.
\newblock In \emph{Proceedings of the 2021 Conference of the North American Chapter of the Association for Computational Linguistics: Human Language Technologies}, pages 483--498, Online. Association for Computational Linguistics.

\bibitem[{Zhang et~al.(2021{\natexlab{a}})Zhang, Deng, Li, Yuan, Bing, and Lam}]{zhang-etal-2021-aspect-sentiment}
Wenxuan Zhang, Yang Deng, Xin Li, Yifei Yuan, Lidong Bing, and Wai Lam. 2021{\natexlab{a}}.
\newblock \href {https://doi.org/10.18653/v1/2021.emnlp-main.726} {Aspect sentiment quad prediction as paraphrase generation}.
\newblock In \emph{Proceedings of the 2021 Conference on Empirical Methods in Natural Language Processing}, pages 9209--9219, Online and Punta Cana, Dominican Republic. Association for Computational Linguistics.

\bibitem[{Zhang et~al.(2021{\natexlab{b}})Zhang, Li, Deng, Bing, and Lam}]{zhang-etal-2021-towards-generative}
Wenxuan Zhang, Xin Li, Yang Deng, Lidong Bing, and Wai Lam. 2021{\natexlab{b}}.
\newblock \href {https://doi.org/10.18653/v1/2021.acl-short.64} {Towards generative aspect-based sentiment analysis}.
\newblock In \emph{Proceedings of the 59th Annual Meeting of the Association for Computational Linguistics and the 11th International Joint Conference on Natural Language Processing (Volume 2: Short Papers)}, pages 504--510, Online. Association for Computational Linguistics.

\bibitem[{Zhang et~al.(2022)Zhang, Li, Deng, Bing, and Lam}]{absa}
Wenxuan Zhang, Xin Li, Yang Deng, Lidong Bing, and Wai Lam. 2022.
\newblock A survey on aspect-based sentiment analysis: tasks, methods, and challenges.
\newblock \emph{IEEE Transactions on Knowledge and Data Engineering}.

\end{thebibliography}

\end{document}